\documentclass[10pt, conference, compsocconf]{IEEEtran}
\IEEEoverridecommandlockouts

\usepackage[letterpaper, left=0.63in, right=0.63in, bottom=1in, top=0.75in]{geometry}
\usepackage{cite}
\usepackage{amsmath,amssymb,amsfonts}
\usepackage{algorithmic}
\usepackage{graphicx}
\usepackage{textcomp}
\usepackage{xcolor}
\usepackage{url}
\usepackage{comment}
\usepackage{array}
\usepackage{varwidth}
\usepackage{booktabs}
\usepackage[font=small]{caption}
\usepackage{threeparttable}
\usepackage{subfig}
\usepackage{makecell}
\usepackage[dvipsnames]{xcolor}
\definecolor{cb_orange}{rgb}{1.0,0.51,0.0}
\definecolor{cb_blue}{rgb}{0.22,0.49,0.72}
\definecolor{cb_green}{rgb}{0.3,0.67,0.29}
\definecolor{cb_red}{rgb}{0.89,0.1,0.11}
\definecolor{cb_pink}{rgb}{1, 0, 0.4}

\def\BibTeX{{\rm B\kern-.05em{\sc i\kern-.025em b}\kern-.08em
    T\kern-.1667em\lower.7ex\hbox{E}\kern-.125emX}}
\begin{document}

\title{ Design Exploration for Protection and Cleaning of Solar Panels with Case Studies for Space Missions

\vspace{-0.15in} 
}

\author{\IEEEauthorblockN{Cameron Robinson}
\IEEEauthorblockA{\textit{Dept. of Computer Systems} \textit{Engineering Technology} \\
\textit{Oregon Institute of Technology}\\
Klamath Falls, OR, USA \\
cameron.robinson@oit.edu}
\vspace{-0.35in}
\and

\IEEEauthorblockN{Ganghee Jang}
\IEEEauthorblockA{\textit{Dept. of Computer Systems} \textit{Engineering Technology} \\
\textit{Oregon Institute of Technology}\\
Klamath Falls, OR, USA \\
ganghee.jang@oit.edu}
\vspace{-0.35in}
}

\maketitle

\begin{abstract}

Solar energy is used for many mission-critical applications including space exploration, sensor systems to monitor wildfires, etc. Their operation can be limited or even terminated if solar panels are covered with dust or hit by space debris. To address this issue, we designed panel cleaning mechanisms and tested protective materials. For cleaning mechanisms, we designed and compared a wiper system and a rail system.  For protective materials, we found through collision tests that polycarbonate was very promising, though the most important factor was layering a soft material between the panel's surface and a hard material. In the cleaning system comparisons, the wiper-based system was more efficient than the rail-based system in terms of cost, cleaning speed, and total power consumption.
\end{abstract}

\section{Introduction}
\label{section:introduction}
Electric power generation from renewable sources is the key for the successful operation of mission-critical unmanned equipment, such as, wildfire monitoring stations, satellites, Mars rovers, etc. The performance of systems powered by solar energy are heavily dependent on the conditions of the solar panels. The panels must be able to supply enough power to meet the systems' consumption for any given moment. As emphasized by the Mars lander that lost power due to dust \cite{news_mars_2022}, maintaining the solar panels through cleaning is essential.
 
 The importance of protecting solar panels is even more serious in space.
 Debris in satellite orbits can cause damage \cite{noauthor_space_2024}, and sandstorms on Mars can wear the panels, causing total power loss. 
These events can shorten the time for critical operations, especially in cases like these, where panels cannot be easily repaired or replaced. 

Since the condition of solar panels can affect system performance in many ways, 
protection and cleaning are key to improving long-term performance and preventing financial loss. 
In this work, we designed and tested both protection and cleaning methods for the design of an efficient automated maintenance system for solar panels.  

The contribution of our work can be summarized as follows:
\begin{itemize}
    \item Exploration of design space for extra protective materials. For this purpose, we assume protection requirements at orbits in space. 
    \item Exploration of solar panel cleaning system designs. For this purpose, we performed tests for numerical comparison of two different cleaning systems.
\end{itemize}
\section{Previous work and design issues}
\label{section:prevWork}

Electricity from solar panels is expected to take 20\% of the global energy demand share by 2050. The power loss from dirty solar panels is estimated to be up to 50\% as in \cite{noauthor_scientists_nodate}. Therefore, a significant amount of research on solar panel cleaning has been performed.

Rupin, et al. \cite{rupin_dirt_2023} used a rail-based mechanism and was one of the rare works that measured cleaning time as a performance metric. It took 300 seconds to finish a single cleaning process moving the wiper from one side to the other. 

Habib et al. \cite{habib_automatic_2021} experimented with soft materials to show the usefulness of wiper blades to avoid scratching the panels under varying load conditions, but their discussion on the definition of dirt species was weak. 

There are other types of systems such as using robots \cite{phi_tien_design_2024}, static electricity to repel dust \cite{mestnikov_development_2021}, and clean systems for the large scale arrays of solar panels \cite{al-housani_experimental_2019}. These systems were not considered in this study as our target applications are  small sized autonomous systems such as planet rovers and weather stations.

Another design domain of solar panels we are addressing is protection from direct damage. Damage to solar panels can greatly decrease the efficiency of power generation, and can even cause the panel to fail entirely. Existing protective coatings are often applied during the manufacturing process, and mostly consider resiliency against fire \cite{wu_review_2020, sathyanarayanan_comprehensive_2024}. Design consideration in an extreme environment and exploring multiple design domains for efficient implementation have not been clearly discussed so far. 
\section{Proposed ideas}
\label{section:proposedIdea}

\subsection{Protective coating and collision experiment setup}
\label{subsection:collision}
An environment with heavy collisions, such as space orbits, is a good case to explore to determine extra protective materials. Space debris with diameters larger than 1 cm can carry enormous energy, which can demolish satellites or spaceships, but they are tracked by a radar system to avoid collision. On the other hand, small debris with 1 - 2 mm diameters can still cause immense damage, and are harder to track, which is why we chose them as a baseline metric for testing protective materials. 

\begin{figure}[!t]
    \centering
    \subfloat[][Materials]{\includegraphics[width=0.42\linewidth]{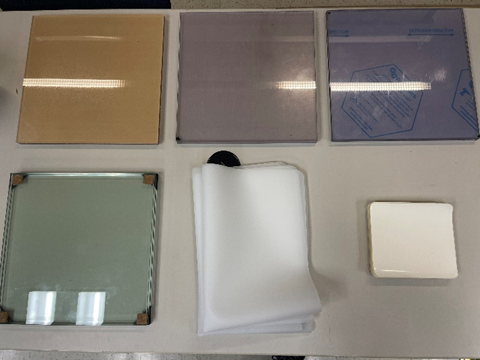}} \hspace{0.05in}
    \subfloat[][Test Setup]{\includegraphics[width=0.25\linewidth]{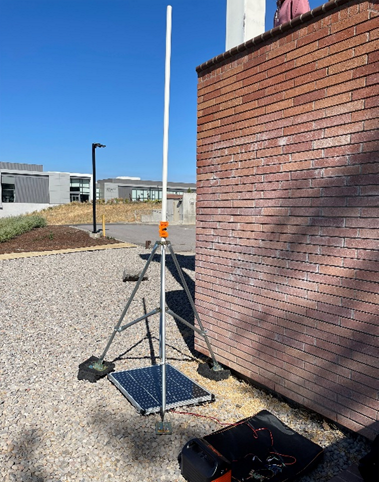}} \hspace{0.05in}
    \caption{Protective Materials and Experiment Setup for Collision Tests}
    \label{fig:ProtectiveMaterials}
    \vspace{-0.19in}
\end{figure}

For our experiments, we designed a 2.3 Kg mass (tip diameter of 1.5 mm) to emulate the collision of debris with 1 - 2 mm diameter. The mass is guided through a PVC pipe mounted in a tripod as shown in Fig. \ref{fig:ProtectiveMaterials}.(b).

The protective materials selected for the collision experiment were polycarbonate (PC), polyethylene terephthalate glycol (PETG), acrylic (AC), laminated glass (LG), silicone (SI), and polyurethane (PU). These materials are all pictured in Fig. \ref{fig:ProtectiveMaterials}.(a). Dimensions and costs of each material are listed in TABLE \ref{tab:material_properties}.

We used three different 100 W solar panels. We designed our cleaning mechanisms around a large single-pane solar panel, and performed collision tests on solar panels with multiple panes (2 - 4 sub-panels summing to 100 W) that could be broken individually. To account for changes in weather and total power rating of the panels being tested, we calculated the percent change in power output by the panels.  

\begin{table}[ht]
\centering
    {
    \begin{tabular}{|l|c|c|c|c|}
        \hline
        \textbf{Material} & \textbf{Thickness (cm)} & \textbf{Area (cm$^2$)} & \textbf{Cost/Area  (\$/cm$^2$)} \\ \hline

        PC & 0.635 & 929 & 0.017\\ \hline
        PETG & 0.635 & 929 & 0.011\\ \hline
        AC & 0.59 & 929 & 0.015 \\ \hline
        LG & 0.635 & 929 & 0.102 \\ \hline
        SI & 0.14 & 2,741.93 & 0.005 \\ \hline
        PU & 0.2 & 234.4 & 0.028 \\ \hline
    \end{tabular}
    }
    \centering
    \caption[]{Material Properties}
    \label{tab:material_properties}
    \vspace{-0.1in}
\end{table}

\begin{figure}[ht!]
    \centering
    \includegraphics[width=.99\columnwidth]{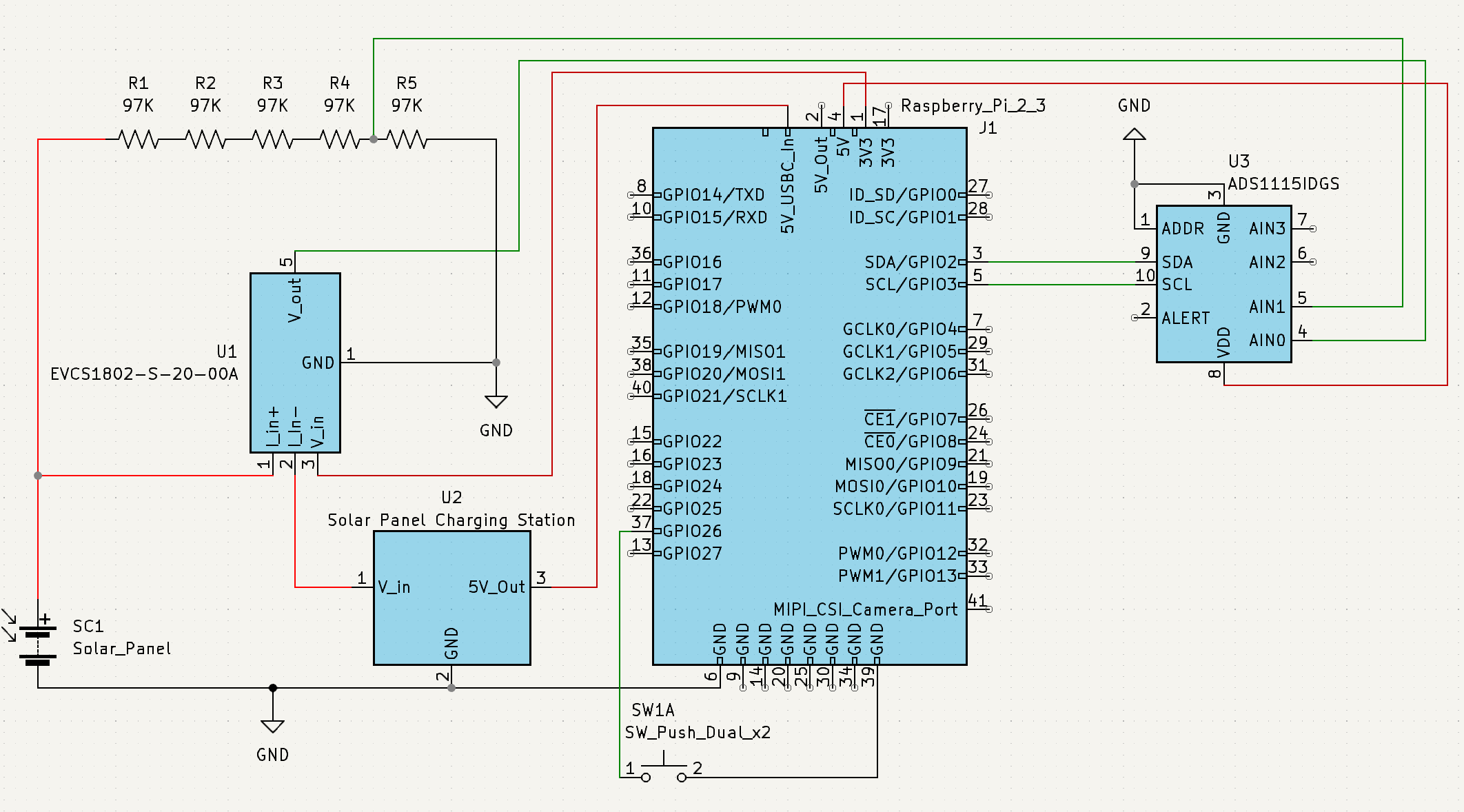}
    \caption{Circuit for Measuring Solar Panel Output}
    \label{fig:CollisionTestCircuit}
    \vspace{-0.19in}
\end{figure}

The measurement system was developed with a Raspberry Pi, analog-to-digital converter (ADC), current sensor, and voltage divider as in Fig. \ref{fig:CollisionTestCircuit}. The voltage divider and current sensor were read into the Raspberry Pi through an analog-to-digital converter. The pushbutton was used to indicate different stages throughout the tests. These stages were before the materials were placed on the panel, after the materials were placed, and after the collision.

\subsection{Cleaning mechanisms: wiper system and rail system}
\label{subsection:mechanism}

In our cleaning mechanism experiment, we implemented two different cleaning systems: a wiper system and a single-axis linear rail system. Both systems used cleaning arms fitted with rubber windshield wiper blades. The solar panels the systems were installed on had dimensions of 91.4 cm x 68.6 cm x 3.56 cm.

TABLE \ref{tab:wiper_parts} lists the parts used to assemble the wiper system, and TABLE \ref{tab:rail_parts} contains the parts used to assemble the rail system. The total cost of parts to build the wiper system was \$648.19, while the total cost of the rail system was \$682.56. In total, the rail system was about 5\% 
more expensive than the wiper system. Each system included custom 3-D printed parts. The wiper system needed one custom part printed eight times for a total of about 96 grams. The rail system needed 
fourteen custom parts weighing a total of 232 grams.

\begin{figure}[htbp!]
    \centering
     \subfloat[Wiper System]{
        \includegraphics[width=0.41\linewidth]{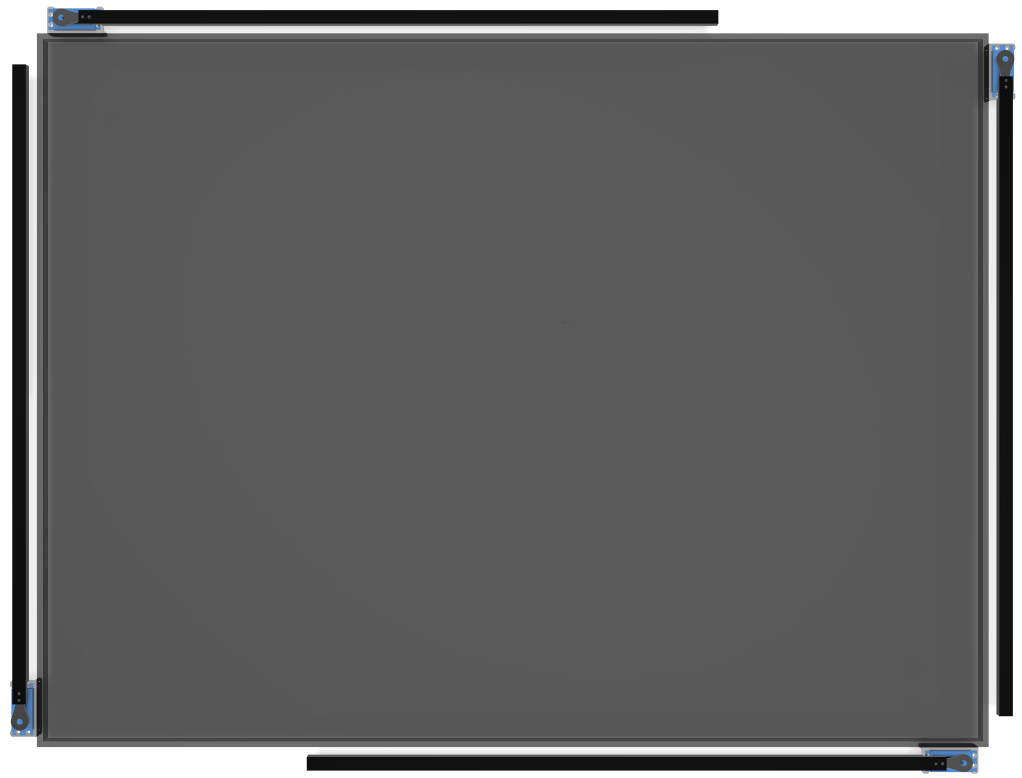}
      }
      \subfloat[Rail System]{
        \includegraphics[width=0.52\linewidth]{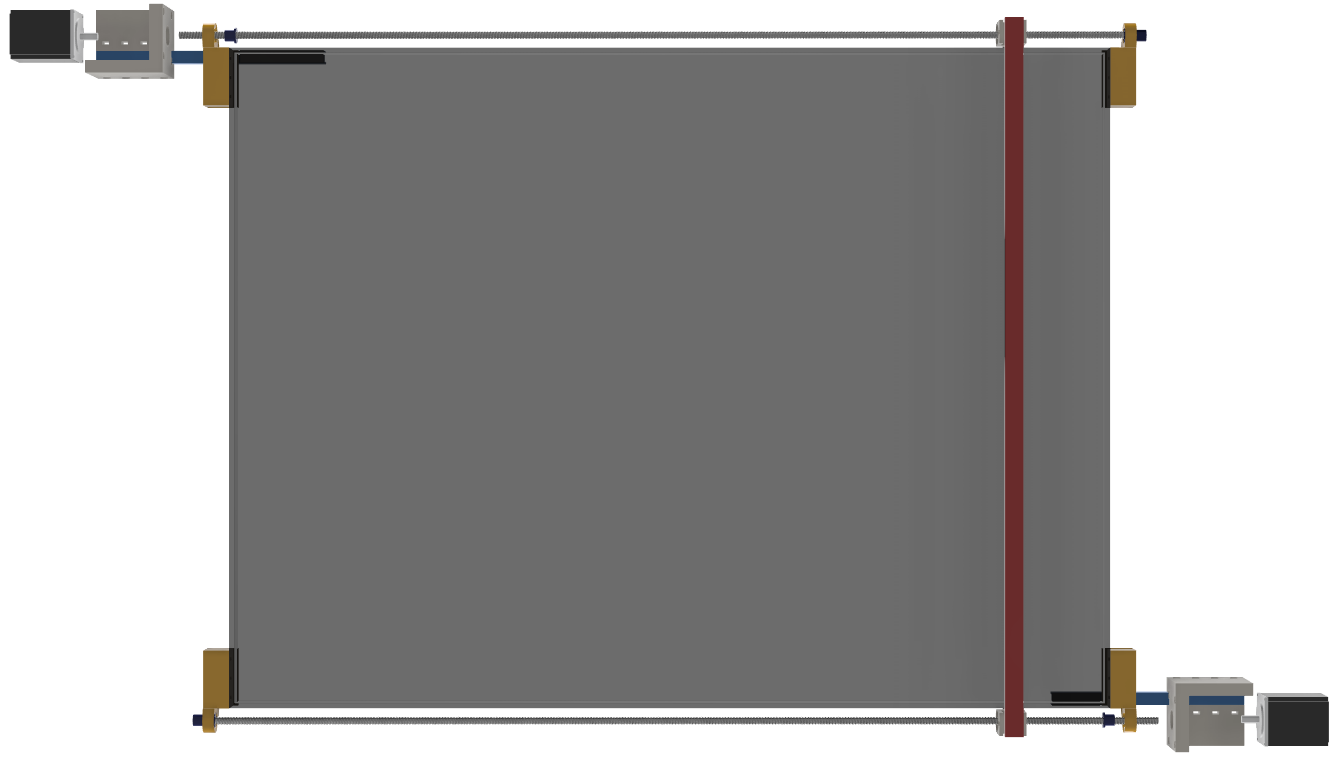}
        }
\caption{
\label{fig:CAD_models}%
Wiper and Rail System CAD Models}
 \vspace{-0.1in}
\end{figure}

For the wiper system, we used four 61 cm cleaning arms, each attached to a servomotor mounted on each corner of the panel as in Fig. \ref{fig:CAD_models}.(a). Each servomotor has a maximum torque of about 3.43 N$\cdot$m, and is operated at 6.6 V. Each arm sweeps 90 degrees over the surface of the panel. We ordered the arm movements to minimize the number of times each motor runs, but also avoid having the wiper arms block dirt from being swept off the panel.

\begin{table}[h]
    \centering
    \begin{tabular}{|l|c|r|}
        \hline
        \textbf{Part} & \textbf{Qty} & \textbf{Cost (USD)} \\
        \hline
        Raspberry Pi 4 & 1 & \$139.99 \\ \hline
        Servo Motor & 4 & \$80.00 \\ \hline
        Solar Panel Charging Station & 1 & \$279.00 \\ \hline
        Current Sensor & 1 & \$10.75 \\ \hline
        32" Rubber Windshield Wiper Blades & 4 & \$7.88 \\ \hline
        16-bit ADC & 1 & \$5.53 \\ \hline
        PETG 3D Printer Filament & 1 & \$2.10 \\ \hline
        5V Buck Converter & 1 & \$42.00 \\ \hline
        Servo Mounting Brackets & 4 & \$22.00 \\ \hline
        48" Aluminum Channel for Wiper Arm & 2 & 28.94 \\ \hline
        Misc. (Screws, wires, resistors, etc.) & 1 & \$30.00 \\
        \hline
         & \textbf{Total:} & \textbf{\$648.19} \\
        \hline
    \end{tabular}
    \caption{Parts List for Wiper System}
    \begin{tablenotes}
    \item[] Note 1: Costs are for entire entry (individual cost * quantity).
    \end{tablenotes}
    \label{tab:wiper_parts}
\end{table}

\begin{table}[h]
    \centering
    \begin{tabular}{|l|c|c|}
        \hline
        \textbf{Part} & \textbf{Qty} & \textbf{Cost (USD)} \\ \hline
        Raspberry Pi 4 & 1 & 139.99 \\ \hline
        NEMA 23 Stepper Motor & 2 & 59.98 \\ \hline
        RPi Stepper Motor HAT & 1 & 22.00 \\ \hline
        Solar Panel Charging Station & 1 & 279.00 \\ \hline
        Current Sensor & 1 & 10.75 \\ \hline
        32" Rubber Windshield Wiper Blades & 2 & 3.94 \\ \hline
        16-bit ADC & 1 & 5.53 \\ \hline
        PETG 3D Printer Filament & 1 & 5.09 \\ \hline
        5V Buck Converter & 1 & 42.00 \\ \hline
        Linear Guide Rails & 2 & 59.88 \\ \hline
        T-Slot Bar 36" & 1 & 17.20 \\ \hline
        Limit Switches & 2 & 1.20 \\ \hline
        6.35mm to 8mm Shaft Coupler & 2 & 11.99 \\ \hline
        Misc. (Screws, wires, resistors, etc.) & 1 & 30.00 \\ \hline
         & \textbf{Total:} & \textbf{682.56} \\ \hline
    \end{tabular}
    \caption{Parts List for Rail System}
    \begin{tablenotes}
    \item[] Note 1: Costs are for entire entry (individual cost * quantity).
    \end{tablenotes}
    \vspace{-0.15in}
    \label{tab:rail_parts}
\end{table}

The rail system has a single cleaning arm attached to two lead screws running along the length of the panel as in Fig. \ref{fig:CAD_models}.(b). The lead screws were rotated using two 12 V NEMA 23 stepper motors with a maximum torque of 1.85 N$\cdot$m and controlled by a Raspberry Pi. A single cleaning cycle moves the arm across the panel once. As the lead of the screw was very dense (2 mm), each cycle took around 200 seconds. If we used a screw with a larger lead, the cleaning speed could be improved. However, since we did not have screws of other sizes, we could not test the system with longer leads.

We selected dry dirt and snow as dirt species that could be found in the planets or orbits in the solar system. 

We measured power consumption until each dirt species was cleaned. The cleaning result was also inspected visually. The dirt species were as follows: 

\begin{itemize}
\item Dry Dirt (soil): 100 grams
\item Snow (icy dirt): 400 grams
\end{itemize}

\section{Results and Discussions}
\label{section:results}
\subsection{Collision Test Results}
\label{subsection:collision_results}

TABLE \ref{tab:collision_results} shows the results of the collision experiments. Materials are listed in the order they were stacked on the solar panel. On their own, the materials did a poor job of protecting the panel. The exceptions were three layers of laminated glass and nine layers of polyurethane. However, while the panel survived in both tests, all three layers of laminated glass shattered and became opaque green, and the solar panel was barely visible beneath the nine layers of polyurethane. Due to the laminated glass being the most expensive material and being useless once shattered, we quickly eliminated it as a potential protective material. The other three hard materials handled impact damage much better since it tended to leave small dents rather than shattering the entire pane.

\begin{table}[ht!]
    \centering
    \begin{tabular}{|c|c|c|}
        \hline
        \textbf{Material Layers} & \textbf{\makecell{Panel\\Broke}}&\textbf{\makecell{Material\\Broke}}\\
        \hline
        3x\footnotemark[1] LG&	N&Y\\
        3x AC&		Y&N \\
        3x PETG&Y&N \\
        3x PC&	Y&N \\
        6x PU&	Y&N\\
        12x SI&	Y&N\\
        9x PU&	        N&N\\
        3x AC, 4x SI&	N&N\\
        4x SI, 3x AC&	Y&N\\
        3x PETG, 4x SI&N&N \\
        3x PC, 4x SI&N&N \\
        3x PC, 9x PU&	N&N \\
        3x PETG, 9x PU	&N&N\\
        3x AC, 9x PU&  N&Y\\
        2x AC, 9x PU&	N&N \\
        2x PC, 9x PU&	N&N \\
        2x PETG, 9x PU&	N&N \\
        2x PETG, 4x SI&	N&N \\
        2x PC, 4x SI&	N&N \\
        2x PC, 2x SI&	Y&N \\
        1x PETG, 4x SI&	Y&Y \\
        1x PC, 4x SI&	N&N \\
        \hline
    \end{tabular}
    \caption{Collision Test Results}
    \label{tab:collision_results}

    \begin{tablenotes}
    \item[] Note 1: multiplier (3x, 6x, ...) indicates the number of stacks of the following material.
    \vspace{-0.15in}
    \end{tablenotes}
\end{table}

We found much more success when testing combinations of materials, particularly when a soft material was placed between the panel and a hard material as a cushion. Layering the soft material on top of the hard material often resulted in the panel breaking (e.g., 3x AC, 4x SI protected the panel, while 4x SI, 3x AC did not). The best combination we found was polycarbonate with silicone under it. The thinnest we got without breaking the panel was one layer of PC, and four layers of SI. While they all performed similarly in terms of protecting the panel, polycarbonate outperformed acrylic, PETG, and laminated glass because it was the only hard material that did not break in any of the tests. For the soft material, silicone performed well and, as shown in TABLE \ref{tab:material_properties}, it was less expensive than polyurethane. 

Throughout our tests, we found that power loss due to protective materials varied between 4\% and 51\% of the measured output of the panels, with an average of about 18\%. Similarly, we found that a shattered solar panel lost anywhere from 20 - 100\% of its power output, with an average power loss of about 43\%. These results indicate that there is value in layering the solar panels with protective materials, though these benefits may vary. Nonetheless, in situations where panels have high chances of shattering, even a 50\% loss in power from materials could be worth it since a single shattering impact, let alone multiple, can cause much more power loss.

\subsection{System Cleaning Test Results}
\label{subsection:cleaning_results}

Throughout the cleaning tests, the wiper system's power consumption ranged from 5.5 to 11.6 W/s, with an average draw of 8.6 W/s. The wiper took about 15 s per cleaning cycle, but needed two cycles to clean the panel, resulting in a total average power consumption of 258 W. The rail system's power consumption ranged from 6.1 to 6.7 W/s, with an average power draw of 6.4 W/s. The rail took only one cycle to clean the panel, but each cycle was about 200 s, resulting in a total average consumption of 1280 W.

Overall, the wiper system runs faster, but has a higher rate of power consumption, while the rail system runs longer, but has a lower, and more consistent, rate of power usage. 

The wiper system performed better for the designed workload. However, the current design setup does not include many dirt species. Therefore, care should be given in the interpretation of the experiment results. 
\section{Conclusions}

Solar panel protection and cleaning system designs were explored in this work. For the development of design protocols, extreme environments such as space orbits and non-Earth planets were considered. 
Our experiments showed that polycarbonate and silicone combinations were effective for protecting the solar panels. The wiper system performed well for cleaning the panels quickly, but at a slight increase to instantaneous power consumption. 
Even with the efforts for the reasonable test cases for  space explorations, there is room for improvement. In future tests, we plan to utilize simulation-based approaches for more accurate modeling.

\section*{Acknowledgment}
This work was partly supported in part through NASA and Oregon Space Grant Consortium, cooperative agreement 80NSSC20M0035 and Oregon Tech Summer Faculty Creativity Grant 2025.
The authors would like to thank students who helped with the experiments and 3D CAD designs. They are Joseph Bayers, Logan Benoy, Coby Feagins, and Chase Uhrich.

\bibliographystyle{ieeetr}
\bibliography{ref}

\end{document}